\documentclass[10pt, a4paper]{article}
\usepackage[utf8]{inputenc}
\usepackage{amsmath, amssymb}
\usepackage{graphicx}
\usepackage{booktabs} 
\usepackage{hyperref} 
\usepackage[margin=1in]{geometry} 
\usepackage{enumitem} 
\usepackage{longtable} 
\usepackage{tcolorbox}
\tcbuselibrary{theorems}
\usepackage{float} 
\usepackage{titlesec}
\titleformat{\section}{\normalfont\Large\bfseries}{\thesection}{1em}{}
\titleformat{\subsection}{\normalfont\large\bfseries}{\thesubsection}{1em}{}
\titleformat{\subsubsection}{\normalfont\normalsize\bfseries}{\thesubsubsection}{1em}{}

\title{From Intent to Execution: Composing Agentic Workflows with Agent Recommendation }
\author{
Kishan Athrey$^{*}$, Ramin Pishehvar$^{*}$, Brian Riordan, and Mahesh Viswanathan \\
Cisco Systems Inc. \\
\texttt{\{kathrey, rpishehv, brriorda, mahviswa\}@cisco.com}
}

\begin{document}
\date{}
\maketitle
\begingroup
\renewcommand\thefootnote{}
\footnotetext{$^{*}$These authors contributed equally.}
\endgroup

\begin{abstract}
Multi-Agent Systems (MAS) built using AI agents fulfill a variety of user intents that may be used to design and build a family of related applications. However, the creation of such MAS currently involves manual composition of the plan, manual selection of appropriate agents, and manual creation of execution graphs. This paper introduces a framework for the automated creation of multi-agent systems which replaces multiple manual steps with an automated framework. The proposed framework consists of software modules and a workflow to orchestrate the requisite task-specific application. The modules include: an LLM-derived planner, a set of tasks described in natural language, a dynamic call graph, an orchestrator for map agents to tasks, and an agent recommender that finds the most suitable agent(s) from local and global agent registries. The agent recommender uses a two-stage information retrieval (IR) system comprising a fast retriever and an LLM-based re-ranker. We implemented a series of experiments exploring the choice of embedders, re-rankers, agent description enrichment, and supervising critique agent. We benchmarked this system end-to-end, evaluating the combination of planning, agent selection, and task completion, with our proposed approach. Our experimental results show that our approach outperforms the state-of-the-art in terms of the recall rate and is more robust and scalable compared to previous approaches. The critique agent holistically reevaluates both agent and tool recommendations against the overall plan. We show that the inclusion of the critique agent further enhances the recall score, proving that the comprehensive review and revision of task-based agent selection is an essential step in building end-to-end multi-agent systems.

\end{abstract}

\section{Introduction}
The landscape of artificial intelligence is rapidly evolving with the increasing availability of specialized agents, both general-purpose and domain-specific. The problem of designing an agentic workflow given a user intent  arises in many areas of AI including workflow automation, multi-agent systems (MAS), embodied AI, etc. These systems involve agents who can plan, reason, and act using tools and memory to achieve a goal. 

We propose AutoMAS to design agentic workflows based on user intent in an automated way and with minimum human intervention. Our proposed approach has fast retrieval, is scalable to a large number of agents in the search space (i.e., agent registry),  is robust to failures through redundancy paths, and uses an optimal path of execution based on user-provided constraints.  Our approach receives a user intent (e.g., create a restaurant reservation app) and optional requirements and constraints from the user or another agent (e.g., optimize for lowest latency at runtime).  It then processes the user intent to generate a plan in natural language consisting of subtasks to be executed in sequence or in recursion. The generation of the plan can optionally leverage  similar previous ones for insight and hence for improved quality. 

In order to account for the non-sequential nature of plans and to include redundant and dynamic paths for robustness and fault tolerance, we represent the plan as a finite state machine (FSM). An Agent-Recommender (AR) maps tasks in the FSM to agents available in local or distributed agent registries. The AR may return more than one match per task to increase robustness and fault tolerance in the workflow (through the creation of alternative routes in the workflow based on the dynamic nature of the environment in which the MAS evolves). While these agents offer  potential for automating complex tasks, their exploding number in the agent search space (i.e., agent registry) presents a new combinatorial challenge: how to efficiently identify the "best" agent for a particular job at a given time. Traditional keyword-based searches, as well as searches using LLMs' input context as the search space, prove inadequate as the agent ecosystem grows. Therefore, we propose the two-stage hybrid search approach that scales better in terms of number of agents in the search space and speed of search. 

More specifically, the design process is as follows:
\begin{itemize}
    \item The user expresses their intent in natural language (e.g., "I want to create a MAS application that makes reservation")
    \item The framework creates a plan in natural language and creates an FSM that corresponds to the generated plan
    \item The framework finds the adequate agent for each node in the FSM and creates an execution graph with inputs and outputs
    \item  At runtime, the end-user expresses their request at the input of the MAS, and the requested task (e.g., "make a reservation at the closest Mexican restaurant") is completed automatically by running the MAS.
\end{itemize}


\section{Background and Motivation}
The core motivation of this work is to automate the process of creating  a MAS, from user intent with minimal human intervention, that can adapt, make decisions, and change its course of actions based on user intent, user requirements, environmental factors, and intermediate results. The generation of a successful MAS at design time requires the successful completion of many steps such as planning, task-to-agent mapping, call graph creation, etc. For the task-to-agent mapping with an Agent Recommender (AR), the difficulty stems from navigating vast agent registries for the optimal set of agents given a task description in natural language and some performance metrics. Furthermore, the challenge intensifies with the scale and diversity of available agents, since we are dealing with a world of agents. An effective recommender system must not only identify relevant agents, but also rank them according to task-specific criteria, constraints, security and safety labels, and application intent. This necessitates a system that goes beyond simple similarity matching, incorporating advanced information retrieval (IR) techniques, and leveraging the capabilities of large language models for enhanced understanding and ranking.

\section{Related Work}

The development of Multi-Agent Systems (MAS) and agentic workflows has evolved rapidly, shifting from simple tool-use to complex, autonomous planning, agent/tool selection, and execution. The transition from single-prompt interactions to complex workflows with adequate execution graph and tool/agent calling  is a cornerstone of modern AI.  

The agentic workflow and the sequencing of agents is usually captured in an execution graph. Execution graphs can be direct acyclic graphs (DAG) (\cite{wei2025beyond}), finite state machines (\cite{metaagent2025}), or any other type of data structure that allows for the sequencing of agent execution. 

When it comes to the mapping of the tasks in the execution graph to actual executable agents, many different approaches are used in the literature. In \cite{wei2025beyond, shen2024taskbench}, the context of an LLM is used to input the list of all available agents, and the decision about the best fit for each task is performed by the LLM itself. In \cite{autotool2025, graphrag2025, anymac2025, gtool2026}, graph-based tool selection based on the conditional probability of selecting an agent given given selected agent(s) is proposed. In this type of approach, the graph that captures order of execution of agents is learned from training data or is updated in an online fashion. 

For evaluation of MAS, frameworks such as TaskBench \cite{shen2024taskbench}  have established benchmarks for evaluating how LLMs decompose user intent into actionable subtasks, without integrating agent recommendation and retrieval techniques proposed here. Research in this area focuses on the ``ReAct'' (Reason + Act) paradigm (among others), where agents iteratively plan and execute steps. However, as noted by Shen et al. \cite{shen2024taskbench}, the structural fidelity of these plans---specifically the ability to maintain correct dependencies and workflow topology---remains a significant challenge that requires structured prediction and evaluation metrics.

As the ecosystem of specialized AI agents grows, identifying the ``best'' agent for a specific task has become a ``needle in an agentic stack'' problem. Traditional keyword-based search is increasingly insufficient. Huang et al. \cite{huang2023metatool} introduced the ToolE (MetaTool) benchmark to evaluate an LLM's ability to decide when to use a tool and which one to select. Recent studies, such as those by Shi et al. \cite{shi2025retrieval}, suggest that standard retrieval models are often not ``tool-savvy,'' struggling with the semantic nuances of tool descriptions. To address this, Re-invoke \cite{reinvoke2024} proposed tool invocation rewriting to improve zero-shot retrieval, while ToolGen \cite{toolgen2025} explores unified retrieval and calling via generative approaches.

The scientific community has largely converged on two-stage retrieval architectures to balance speed and precision. The first stage typically utilizes hybrid search---combining dense vector embeddings with sparse keyword matching. Hsu et al. \cite{hsu2025dat} demonstrated the efficacy of Dynamic Alpha Tuning (DAT) for optimizing this hybrid balance in Retrieval-Augmented Generation (RAG) systems. The second stage involves LLM-based re-rankers, which provide a more nuanced semantic understanding of the task-agent fit. Our work builds on these foundations by incorporating agent description enrichment, a technique supported by Chen et al. \cite{reinvoke2024}, to enhance the semantic representation of agents at ingestion time.

Finally, the concept of ``self-correction'' or ``critique'' is gaining traction as a method to improve the reliability of MAS. Lim et al. \cite{itertool2024} introduced ITR-RAG, which uses iterative tuning and feedback from LLMs to refine retrieval results. Similarly, Wang et al. \cite{reagt2025} proposed Reagt, focusing on trustworthiness in knowledge-intensive tasks. While these methods often focus on the accuracy of the retrieved information, our framework extends the critique mechanism to evaluate the global optimality of an agent within a complex, multi-step workflow, ensuring input-output compatibility and adherence to user-defined constraints like cost and latency.

\section{Architecture of AutoMAS}
\begin{figure}[h!]
    \centering
     \includegraphics[width=1\textwidth]{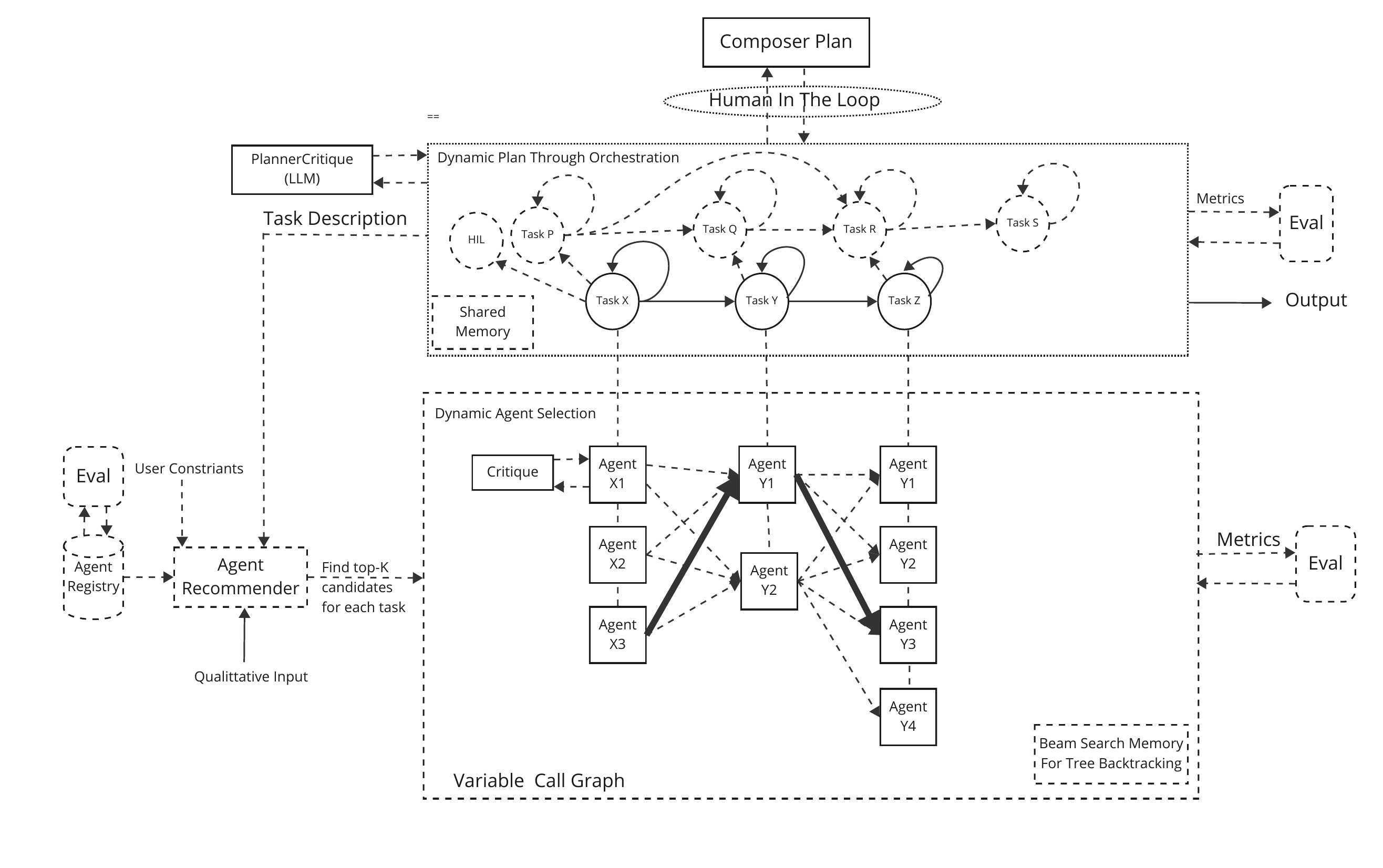} 
    \caption{Architecture for an end-to-end MAS with dynamic and redundant workflow}
    \label{fig:architecture_overview}
\end{figure}

We propose a MAS architecture that takes user intent and requirements to generate a flexible plan that can dynamically change based on the environmental feedback and human-in-the-loop. The dynamic plan maps to a call graph with multiple routes to increase redundancy and robustness. The mapping between the tasks in the plan and agents in the call graph is done via the Agent Recommender (AR) by retrieving adequate agents from the Agent Registry  (see section \ref{sec:AR}). The workflow can use different routes (based on the different candidates returned by the agent recommender) via the Variable Call Graph (VCG). The routes themselves may be dynamically selected based on state of the network and compute nodes (e.g., agent X being unavailable due to a network shutdown), or may be optimized based on constraints (e.g., lowest possible latency) as shown in Figure \ref{fig:architecture_overview}).

\subsection{Planner}\label{sec:Planner}
The planner receives a user intent (i.e., a request for the completion of a task from a user in plain natural language) as input and generates at its output a list of subtasks that will run either in sequence or in an finite state machine. Planning based on user intent can be done using an LLM. The output of this stage is an FSM with nodes that identify tasks in natural language. The planner can also leverage a database of previously designed plans (episodic and semantic memories) or a dataset of publicly available plans to refine the generated plan (i.e., retrieval augmented generation with a knowledge base of previous plans). 

\subsection{Variable Call Graph}
The Variable Call Graph (VCG) is a rerouting mechanism as depicted in Fig. \ref{fig:architecture_overview}, in which the different routes are generated based on the different candidates returned by the agent recommender (as explained in section \ref{sec:AR}). It allows for the adjustment of the call graph based on dynamically changing cost of execution, latency, and other environmental factors. To showcase the concept of VCG we ran the following experiment, in which we demonstrate how the variable call graph can optimize the agent calling based on some user-provided optimization constraints. 
\begin{itemize}
\item \underline{Example user intent}: "I am located in Downtown Los Angeles, and I want to find the best Mexican restaurant near me and I'd like to make a reservation"
\item \underline{Generated Plan by LLM}: 1) Search for the nearest Mexican restaurant, 2) Gather and analyze reviews, 3) Make a reservation. 
\item \underline{Retrieved agent-based performance}: For each retrieved agent, cost of execution, speed of execution, and accuracy are retrieved as shown in Table \ref{table_1}
\item \underline{User constraints}: "Pick agents for the tasks that provide the best trade-off between cost and reservation time (but not accuracy)". The result of the agent selection is shown in Box \ref{box:agents}.
\end{itemize} 

\newcounter{mybox}[section]
\renewcommand{\themybox}{\thesection.\arabic{mybox}}

\newtcolorbox{mybox}[2][]{%
  colback=gray!5,
  colframe=black,
  title={Box~\themybox: #2},
  #1
}

\refstepcounter{mybox}\label{box:agents}
\begin{mybox}{Agent Selection Based on User Constraints}

\begin{verbatim}
We prioritize cost and speed over accuracy for this task.

### Selected Agents
1. **Task 1: Search for the Nearest Mexican Restaurant**
   - **Agent Map 1**: [10 cents, 100 ms, 0.95 accuracy]
     - **Reason**: Lowest cost and reasonable speed with high accuracy.

2. **Task 2: Gather and Analyze Reviews**
   - **Agent Review 1**: [10 cents, 20 ms, 0.9 accuracy]
     - **Reason**: Lowest cost and fastest speed with decent accuracy.

3. **Task 3: Make a Reservation**
   - **Phone Call Agent**: [10 cents, 20 ms, 0.7 accuracy]
     - **Reason**: Lowest cost and fastest speed, despite lower accuracy.
\end{verbatim}
\end{mybox}

\begin{table}[h]
\centering
\scriptsize 
\begin{tabular}{l c c c c c c c c r}
\toprule
Metrics &MA/V1 & MA/V2 & MA/V3 & RA/V1  & RA/V2  & RA/V3 & RS(P) & RS(I) & RS(E) \\
\midrule
Cost of Operation (cents) &10 &20 &30 &10 &20 &50 &10 &20 &50 \\
Speed of Execution (ms) &100 &200 &50 &20 &50 &500 &20 &100 &200 \\
Accuracy &0.95 &.90 &.80 &0.90 &0.95 &0.99 &0.70 &0.99 &0.96 \\
\bottomrule
\end{tabular}

\caption{List of agents for the different tasks generated by the plan.  
MA: Mapping Agent, RA: Review Agent, RS: Reservation Agent.  
V1: Vendor 1, V2: Vendor 2, V3: Vendor 3.  
P: Phone, I: Internet, E: Email.  
For each agent we retrieve three performance metrics in the following order: cost of operation (in cents), speed of execution (in milliseconds), and accuracy (0 to 1).}
\label{table_1}
\end{table}

\leavevmode
\subsection{Human-in-the-Loop}
Our proposed network is capable of interacting with a human at design time to improve the multi-agent application, as well as at runtime to get feedback from a human in the case of uncertainty, or when more information is required due to ambiguity.  The workflow also has the ability to be interrupted by a human supervisor through direct inclusion of human-in-the-loop nodes in the execution graph, for the case when wrong behavior is detected by the supervising human.

\subsection{Agent Recommender} \label{sec:AR}
In order for the MAS to function as expected, we need a proper mapping between tasks generated by the planner (as described in section \ref{sec:Planner}) and agents available to the framework. In order to do so, an agent recommender matches the description of agents from a list of agents (i.e., agent registry) to the subtask in the FSM to execute and consequently selects the most optimal agent for the subtask. 

Each agent in the directory is represented by attributes such as agent-ID, name, description in natural language, performance metrics (e.g., accuracy, latency, cost of execution in dollars) and constraints. The schema of the agent registry is governed by a contract between the agent vendors and the users of the system. The agent registry can be generated automatically by GenAI using the documentation associated with the agentic software provided by the vendor.

\begin{figure}[h!]
    \centering
     \includegraphics[width=1.1\textwidth]{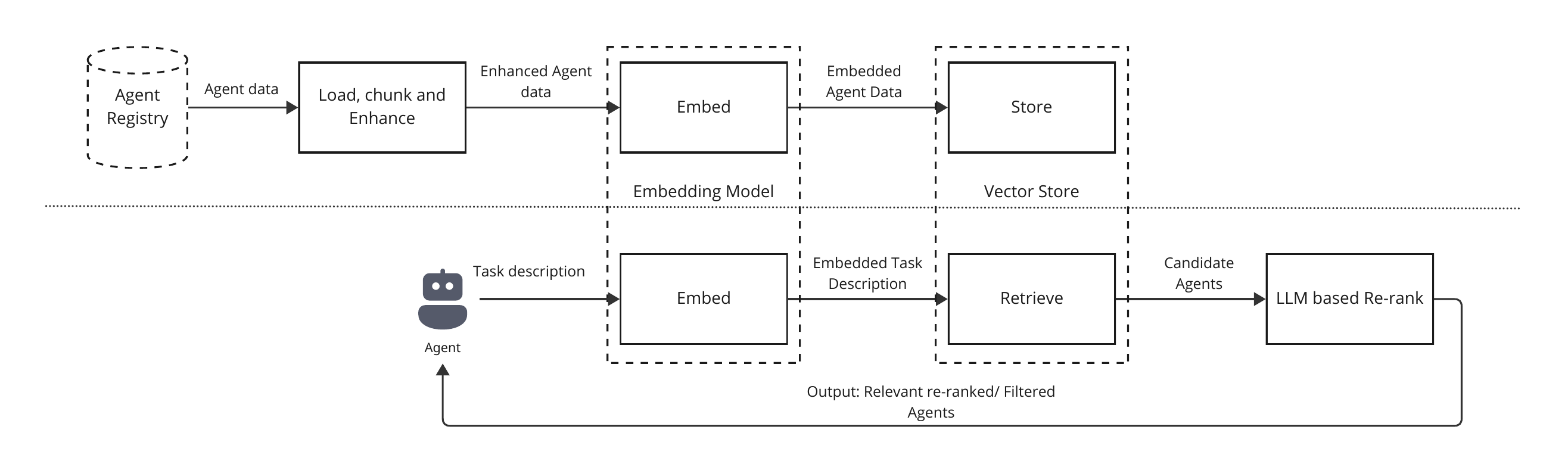} 
    \caption{Agent Recommender that includes the enrich, retrieve, and re-rank phases.}
    \label{fig:AR_recommender}
\end{figure}

The recommender consists of a two-stage retrieval module (that includes a retriever and a re-ranker), as well as a agent description enrichment module \ref{sec:enrichment} as explained in the following sub-sections (see also figure \ref{fig:AR_recommender} for more details).

\subsubsection{Two-Stage Retrieval Process} \label{sec:2stage-retriever}
The core of the recommender as depicted in Fig. \ref{fig:AR_recommender} is a two-stage process:
\begin{enumerate}[label=(\roman*)]
    \item \textbf{Stage 1: Retriever:} This stage employs a fast retrieval mechanism, utilizing hybrid search (keyword + embeddings). Its primary goal is to quickly identify a broad set of potentially relevant agents (high recall). The retriever accounts for constraints specified by multi-agent application developers.
    \item \textbf{Stage 2: Re-ranker:} The candidates identified by the retriever are then passed to a re-ranker. This stage leverages LLMs to re-evaluate and re-order the retrieved agents based on a better semantic understanding of the task and its context, aiming to significantly boost precision, especially at the top of the results list.
\end{enumerate}
The core recommender's performance is further improved by ancillary modules such as agent description  enrichment and critique mechanism as described below.

\subsubsection{Agent Description Enrichment} \label{sec:enrichment}
\begin{figure}[h!]
    \centering
     \includegraphics[width=1.05\textwidth]{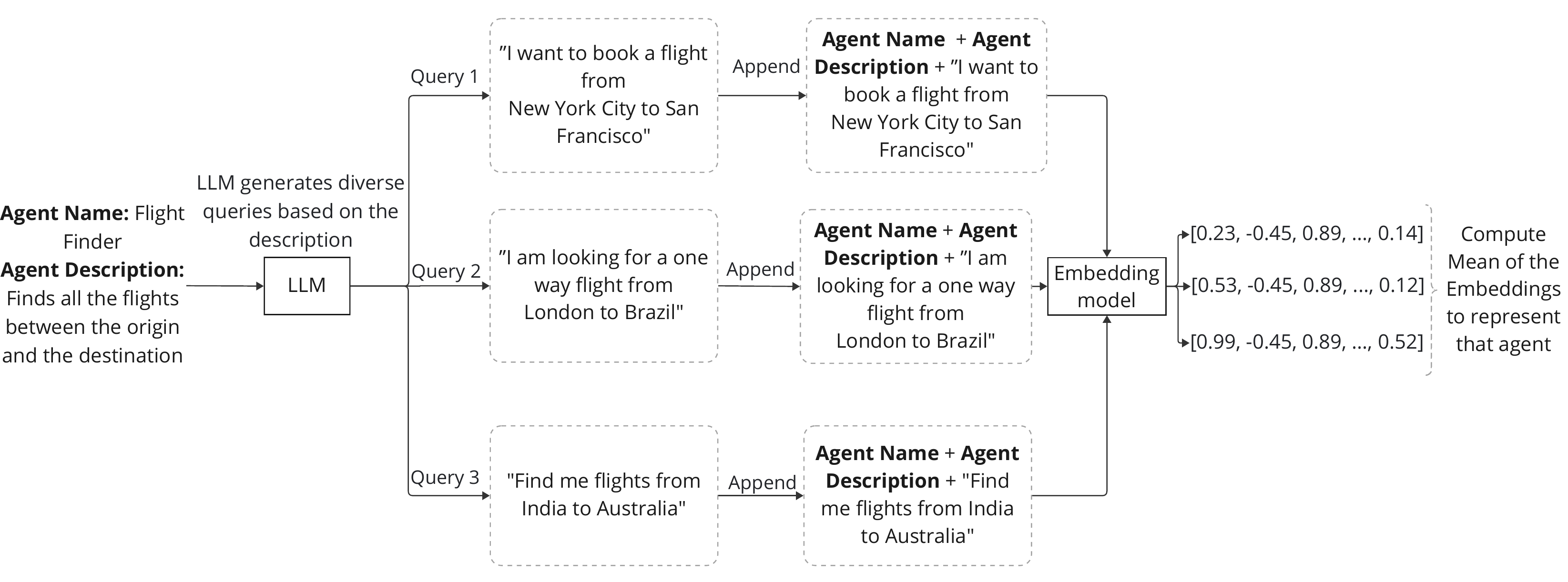} 
    \caption{Agent description enrichment by appending synthetic queries generated from the agent description.}
    \label{fig:Agent Description Enrichment}
\end{figure}

This module focuses on enriching agents' metadata to improve the retrieval quality (see also \cite{reinvoke2024}) of the two-stage retriever described in section \ref{sec:2stage-retriever}. Enriching an agent's metadata involves generating a fixed number of synthetic queries from the agent description field in the agent registry (see Figure \ref{fig:Agent Description Enrichment}).  These synthetic queries, combined with the name and description, form an "enriched document", that is used to search and retrieve agents. For example, if an agent's description is "find all the flights between the origin and the destination", adding synthesized queries like "I want to book a flight from New York City to San Francisco", and appending it to the description at ingestion time improves agent recommendation. Embeddings of these enriched agent descriptions (documents) are averaged to create a single, more robust embedding vector per agent, enhancing the semantic representation for retrieval. Note that agent description enrichment occurs only once at ingestion time (i.e., indexing) and thus does not increase the complexity at retrieval time.

\subsubsection{Critique Mechanism}
The two-stage agent recommender finds the best matching agent for a given subtask without considering the overall fitness of the retrieved agent within the MAS. In other words, it does not take into account additional criteria such as constraints specified by the user (latency, cost, security, safety, etc.), the overall fit of the recommended agents in terms of input-output compatibility with other agents in the VCG, or its optimality given the overall user intent and requirements (i.e., global optimality vs. local optimality).  An optional third stage, the critique, is applied to further refine the task-agent assignment. The critique takes into account the aforementioned aspects with a view to optimizing across the entire function of the MAS. The critique does not re-rank agents but prepares recommendations for consumption by a planner (manual, human-based, or automated). The critic mechanism operates in two modes:
\begin{itemize}
    \item \textbf{In-the-loop (task-specific):} The critique has access only to the current task description and the retrieved and re-ranked agents. It evaluates if the recommendations by the two-stage retriever are adequate or not. It then either proposes better fitting agents or suggests refining the task plan (if adequate agents are not available). This first mode is the less complex implementation of the critique.
    \item \textbf{Outside-the-loop (composition/plan-specific):} In this mode, the critique is provided with a broader context, which includes the overall workflow, task plan, and task graph. This allows for an evaluation of the adequacy of the recommendations of the two-stage retriever based on the entire application's requirements, not just the constituent tasks. This is the more complex implementation of the critique. 
\end{itemize}

\section{Experimental Setup and Results}
We developed  a series of models with increasing complexity. For each component, we determine the optimal hyperparameters and benchmark the system end-to-end. We first performed experiments related to agent recommendation using the ToolE dataset \cite{huang2023metatool}, comprising 199 agents (tools) and 20,550 query-tool pairs for evaluation. The diversity and breadth of the ToolE dataset gave us the opportunity to optimize the agent recommendation pipeline and report the best performing setup without relying on smaller datasets that would have generated statistically noisy results. In what follows, we first report the performance metrics including nDCG (normalized Discounted Cumulative Gain), recall, and mAP (mean Average Precision), and identify the best performing configuration for AR. However, the ToolE dataset is not designed for evaluating planning, since proper breakdown of overall task/user intent into smaller tasks is not included in the dataset. Therefore, we also evaluate the planning performance using a smaller public dataset that includes subtasks (i.e., TaskBench \cite{shen2024taskbench}). We finally report the performance on the overall end-to-end composition task by leveraging the best configuration from the AR experiments conducted with ToolE dataset on our designed planning and composition experiment, which uses TaskBench.  

\subsection{Dataset and Augmentation}
The ToolE dataset was augmented by generating `N` instances (N being variable as described in the `number of synthesize queries` in Table \ref{tab:doc_enrichment}) of the same description with synthetic queries using an LLM as described in the agent description enrichment section \ref{sec:enrichment}. Average embeddings were computed for each instance to enhance document representation.

\subsection{Impact of Embedders}
We compared different embedding models with and without re-ranking by using different metrics on the ToolE dataset and reported the results  in Table \ref{tab:embedder_comparison}, with the "retriever only" row being the baseline. Throughout this article, we used a hybrid alpha search for retrieval with Weaviate \cite{weaviate} (when the tuning parameter "hybrid value" of 0.9 was used as a trade-off parameter between dense and sparse search methods as described in \cite{hsu2025dat}. As expected,  larger embeddings generally improve recall, particularly in the "retriever only" setup. Even though the difference in precision (nDCG@1) was more pronounced for the retriever alone, this gap significantly narrowed once a re-ranker was applied. 

\begin{table}[h!]
    \centering
    \caption{Effect of choice of embedding models (Hybrid value = 0.9, Re-ranker = gpt-4o)}
    \label{tab:embedder_comparison}
    \begin{tabular}{llcccccc}
        \toprule
        Method & Embedding & nDCG@1 & nDCG@5 & Recall@1 & Recall@5 & mAP \\
        \midrule
        Retriever only & text-embedding-3-small & 0.5794 & 0.7166 & 0.5793 & 0.8306 & 0.6783 \\
        Retriever + re-ranker & text-embedding-3-small & 0.7105 & 0.7628 & 0.7104 & 0.7955 & 0.7513 \\
        Retriever only & text-embedding-3-large & 0.6311 & 0.7622 & 0.6313 & 0.8691 & 0.7262 \\
        Retriever + re-ranker & text-embedding-3-large & 0.7355 & 0.7888 & 0.7333 & 0.8233 & 0.7769 \\
        \bottomrule
    \end{tabular}
\end{table}

\subsection{Impact of Re-rankers}
The re-ranker consistently provided a substantial boost in early precision, as detailed in Table \ref{tab:reranker_comparison}. With "text-embedding-3-large" \cite{openlarge}, adding a re-ranker (e.g., `o1` or `gpt-4o`) significantly increased `nDCG@1`. However, this gain in precision came at a slight cost to breadth, with `Recall@5` dropping, which indicates that the re-ranker prioritizes the most relevant items at the top.

\begin{table}[h!]
    \centering
    \caption{Effect of choice of re-ranker models (Hybrid value = 0.9, Embedding = text-embedding-3-large)}
    \label{tab:reranker_comparison}
    \begin{tabular}{llcccccc}
        \toprule
        Method & Re-Ranker & nDCG@1 & nDCG@5 & Recall@1 & Recall@5 & mAP \\
        \midrule
        Retriever only & -- & 0.6307 & 0.7620 & 0.6306 & 0.8689 & 0.7260 \\
        Retriever + re-ranker & o1 & 0.7324 & 0.7891 & 0.7323 & 0.8243 & 0.7767 \\
        Retriever only & -- & 0.6309 & 0.7622 & 0.6309 & 0.8691 & 0.7262 \\ 
        Retriever + re-ranker & gpt-4o & 0.7355 & 0.7888 & 0.7333 & 0.8233 & 0.7769 \\
        \bottomrule
    \end{tabular}
\end{table}

\subsection{Impact of Critique}
The critique validates the choice of the agent from the two-stage retriever against one or more of the following: user intent, user requirements/constraints, call graph, and generated plan. It can optionally work in concert with the planner and the two-stage retriever to refine the output of either modules to improve results. Applying the critique mechanism when top-1 agent is returned  per task demonstrated further improvements, as shown in Table \ref{tab:critic_performance}. For a `Retriever+Reranker` baseline, the addition of critique boosted both `nDCG@1` and `Recall@1`, suggesting the critique effectively refines selections.

\begin{table}[h!]
    \centering
    \caption{Critique Performance for Mode 1 (top-1 Agent)}
    \label{tab:critic_performance}
    \begin{tabular}{lcc}
        \toprule
        Method & nDCG@1 & Recall@1 \\
        \midrule
        Retriever+Reranker & 70.0\% & 70.0\% \\
        Retriever+Reranker+Critique & 73.5\% & 73.5\% \\
        \bottomrule
    \end{tabular}
\end{table}

\subsection{Impact of Agent Description Enrichment}
Agent description enrichment, particularly when more enriched documents are averaged, consistently improved retrieval performance. Table \ref{tab:doc_enrichment} illustrates that increasing the number of averaged enriched embeddings led to incremental gains in `Recall` and `mAP`.

In addition to re-ranking, we explore a \emph{filtering} variant in which the LLM-based re-ranker is instructed to discard candidates it deems irrelevant. Concretely, when filtering is enabled the re-ranker returns only the subset of retrieved agents that it judges to be relevant to the query, effectively pruning false positives from the candidate set. As shown in Table \ref{tab:doc_enrichment}, combining agent description enrichment with filtering yields the largest gains in both Recall and mAP.

\begin{table}[h!]
    \centering
    \caption{AR – Agent Description Enrichment experiments (Re-ranker = gpt-4o). The number of synthesized queries is the augmentation described in section \ref{sec:enrichment}}
    \label{tab:doc_enrichment}
    \begin{tabular}{llccccc}
        \toprule
        Method & Retrieval k & \# of synthesized queries & Recall & mAP \\
        \midrule
        Retriever + re-ranker & 5 & 3 & 0.8004 & 0.7866 \\
        Retriever + re-ranker & 5 & 5 & 0.8043 & 0.7903 \\
        Retriever + re-ranker & 5 & 10 & 0.8058 & 0.7919 \\
        Retriever + re-ranker & 10 & 10 & 0.8122 & 0.7963 \\
        Retriever + re-ranker & 5 & 10 (with filtering) & 0.8346 & 0.8108 \\ 
        Retriever + re-ranker & 10 & 10 (with filtering) & 0.8398 & 0.8173 \\ 
        \bottomrule
    \end{tabular}
\end{table}

\subsection{Overall Performance and Comparison of Agent Recommendation}

Results in Table \ref{tab:best_methods_comparison} show that our best configuration, incorporating a re-ranker with $k{=}10$ and agent description enrichment with 10 averaged embeddings combined with filtering, significantly outperformed both the baseline and the Re-Invoke approach. A key insight from the non-filtering experiments is that re-ranking alone consistently improved ranking quality (e.g., mAP from 0.734 to 0.805, nDCG@1 from 0.632 to 0.728) without compromising coverage (Recall@10 remained high), while adding filtering further boosted precision by removing irrelevant candidates from the final result set.

\begin{table}[h!]
    \centering
    \caption{Our best methods comparison with Re-invoke paper (Hybrid value = 0.9)}
    \label{tab:best_methods_comparison}
    \begin{tabular}{llccccc}
        \toprule
        Method & Re-Ranker & nDCG & Recall & mAP \\
        \midrule
        Ours Retriever + re-ranker (k=5, 10 avg emb) & gpt4o & 0.7427 & 0.8058 & 0.7919 \\
        Ours Retriever + re-ranker (k=5, 10 avg emb, w/ filtering) & gpt4o & 0.7427 & 0.8346 & 0.8108 \\
        Ours Retriever + re-ranker (k=10, 10 avg emb) & gpt4o & 0.7416 & 0.8122 & 0.7963 \\
        Ours Retriever + re-ranker (k=10, 10 avg emb, w/ filtering) & gpt4o & 0.7416 & 0.8398 & 0.8173 \\
        \midrule
        Re-Invoke w/ BM25 & gpt-3.5 turbo & 0.5255 & 0.6300 & 0.5255 \\
        Re-Invoke w/ Vertex AI & text-bison@001 & 0.6716 & 0.7821 & 0.6715 \\
        \bottomrule
    \end{tabular}
\end{table}

\section{End-to-End Performance: Putting it all together}
\begin{figure}[h!]
    \centering
    \includegraphics[width=0.7\textwidth]{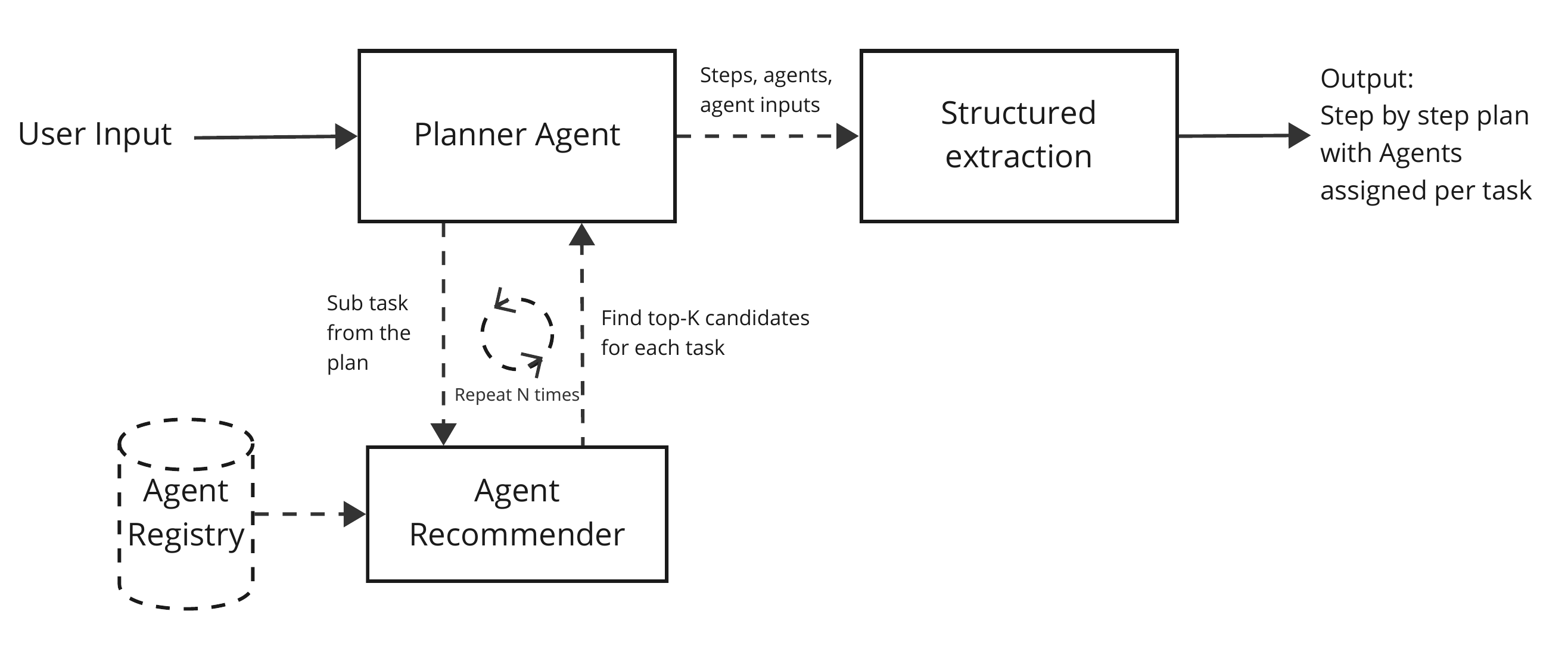} 
    \caption{ReAct Style Architecture \cite{yao2022react} For Planning/Composition. The Planner generates sub tasks and iteratively queries Agent Recommender for each sub task. The framework assigns the suitable tool/agent to each subtask and repeats it. It completes execution when all subtasks are finished. The extractor node collects this information and formats it in the required output format.}
    \label{fig:react_overview}
\end{figure}

We created a model that includes all building blocks described in previous sections (i.e., agent-recommender and planner). We built an end-to-end system that takes a “task” (or intent) as input, decomposes the task into sub tasks, iteratively assigns a tool/agent for each sub task by repeatedly querying the agent recommender and produces a task plan with tool/agent assignment (i.e., ReAct \cite{yao2022react}  for planning/composition) as seen in Figure \ref{fig:react_overview}. We evaluated this approach on the TaskBench dataset.

More specifically, we used the TaskBench with the following composition with a combined total of 103 tools/agents and 17000 queries: 
\begin{itemize}
    \item Dailylife APIs: 40 tools/agents and 4800 queries
    \item Hugging Face APIs: 23 tools/agents and 7500 queries
    \item Multimedia APIs: 40 tools/agents and 5555 queries
\end{itemize}

The results can be found in Table \ref{table:table_BERT}.

\begin{table}[h!]
\centering
\caption{Performance Results with different semantic-based metrics (Rouge and BERTScore), all results are averaged over all samples available in the specified experiment.} \label{table:table_BERT}
\begin{tabular}{lcccccc}
\toprule
\textbf{Category} & \textbf{Rouge 1} & \textbf{Rouge 2} & \textbf{Rouge L} & \textbf{BERT Precision} & \textbf{BERT Recall} & \textbf{ BERT F1} \\

dailylifeapis & 0.603 & 0.354 & 0.504 & 0.519 & 0.551 & 0.535 \\

huggingface & 0.49 & 0.245 & 0.422 & 0.397 & 0.631 & 0.487 \\

multimedia & 0.512 & 0.277 & 0.483 & 0.357 & 0.642 & 0.458 \\
\midrule
combined & 0.531 & 0.298 & 0.479 & 0.421 & 0.592 & 0.492 \\
\bottomrule
\end{tabular}
\end{table}

However, semantic-based scores bring challenges when it comes to tool calling and planning. For instance, the planner may decide to break subtasks further compared to the "oracle" plan or may decide to merge some subtasks. In both cases, both the BERTScore and the Rouge scores will score low, even though the task would be completed as expected. Furthermore, the tool calling can use different filler words compared to the oracle, which are  not detrimental to the execution of the overall plan. Therefore, we created a metric based on LLM-as-a-judge to overcome the aforementioned shortcomings in evaluation.

\subsection{LLM-as-a-Judge Evaluation}
To address the limitations of token-level similarity metrics, we employ an LLM-based evaluation paradigm that assesses task plans through semantic understanding rather than surface-level text matching. The evaluator receives the original user query, the predicted task sequence, and the ground truth task sequence as inputs. The assessment proceeds along two complementary dimensions.

The first dimension, \textit{Overall Match}, determines whether the predicted plan holistically aligns with the reference plan. The evaluator considers semantic equivalence rather than lexical identity. In other words, the predicted plan does not need to share identical wording, but must convey the same intent, invoke the same tools or agents, and specify consistent parameters. A prediction receives a ``Match'' designation only when all constituent steps align with their ground truth counterparts in both meaning and ordering; any deviation in step content or sequence yields a ``no\_match'' verdict.

The second dimension, \textit{Step-Wise Ratio}, provides a more granular assessment by comparing predictions and ground truth in a position-by-position manner. For each index in the ground truth sequence, the corresponding predicted step is evaluated for semantic equivalence. The ratio of matched positions to total ground truth steps produces a continuous score between 0.0 and 1.0, capturing partial correctness when full alignment is not achieved. This metric proves particularly valuable for understanding where plans diverge and quantifying the degree of agreement even in cases of overall mismatch.

Table \ref{tab:performance_metrics} presents the results obtained using this evaluation methodology across the three TaskBench subsets. The \textit{dailylifeapis} category exhibits the strongest performance with a mean overall match rate of 65\% and a step-wise ratio of 0.805, indicating that even when plans do not perfectly match, substantial agreement exists at the individual step level. The \textit{huggingface} subset proves most challenging, likely due to the specialized nature of machine learning pipeline construction. (\textit{multimedia} lands in the middle.) Across all categories combined, the system achieves a 59.3\% overall match rate with an average step-wise correspondence of 0.742, demonstrating that the planner-recommender pipeline generates semantically appropriate task decompositions for a majority of queries.

\begin{table}[h!]
    \centering
    \caption{Performance Metrics by Category with LLM-as-a-judge as the performance evaluator}
    \label{tab:performance_metrics}
    \begin{tabular}{lcc}
        \toprule
        \textbf{Category} & \textbf{Mean Overall Match} & \textbf{Mean Step Wise Ratio} \\
        \midrule
        dailylifeapis & 0.650 & 0.805 \\
        huggingface & 0.527 & 0.676 \\
        multimedia & 0.586 & 0.723 \\
        \midrule
        Combined (all) & 0.593 & 0.742 \\
        \bottomrule
    \end{tabular}
\end{table}

\subsection{Structured Prediction Evaluation}
Beyond semantic evaluation, we adopt the structured prediction framework introduced in TaskBench \cite{shen2024taskbench} to enable fine-grained analysis of plan quality. This approach treats task planning as a structured output problem, where the system must predict not only the sequence of tools/agents but also their interconnections and the overall workflow topology. The evaluation decomposes into several orthogonal dimensions, each capturing a distinct aspect of plan correctness.

\textit{Tool/Node Name Metrics} assess the accuracy of tool selection independent of ordering. Precision measures the fraction of predicted tools that appear in the ground truth, while recall quantifies the coverage of ground truth tools in the prediction. The F1-score provides a balanced aggregate of these complementary measures.

\textit{Argument Metrics} evaluate parameter specification through the Task-ArgName F1 score, which measures whether the predicted argument names for each task align with the expected parameters. This metric proves particularly discriminative, as correct tool identification does not guarantee proper parameterization.

\textit{Sequence Metrics} capture the structural fidelity of predicted plans. Edit distance (normalized Levenshtein distance) quantifies the minimum number of insertions, deletions, and substitutions required to transform the predicted sequence into the ground truth, with lower values indicating better alignment. Sequence similarity provides the complementary view, measuring the degree of correspondence between orderings.

\textit{Task Type and Count Metrics} assess higher-level plan characteristics. Type accuracy evaluates whether the predicted workflow belongs to the correct category (e.g., single, chain, or dag), while N-Tools accuracy measures whether the predicted number of tools matches the ground truth count.

Table \ref{tab:exp1_2} presents results across all structured metrics. The \textit{dailylifeapis} category achieves the highest tool identification performance (F1 of 0.9064) and sequence similarity (0.8956), reflecting the relatively straightforward nature of everyday task automation. Argument prediction proves challenging across all categories, with \textit{huggingface} and \textit{multimedia} achieving Task-ArgName F1 scores below 0.45. 

\begin{table}[h!]
    \centering
    \caption{Metrics for ReAct Style with Structured Prediction. $\uparrow$ higher is better, $\downarrow$ lower is better.}
    \label{tab:exp1_2}
    \begin{tabular}{lccc}
        \toprule
        \textbf{Metric} & \textbf{Dailylife APIs} & \textbf{Hugging Face APIs}& \textbf{Multimedia APIs}\\
        \midrule
        \multicolumn{4}{l}{\textit{\textbf{Tool/Node Name Metrics}}} \\
        Precision & 0.9442 & 0.7422 & 0.9031 \\
        Recall & 0.8716 & 0.6274 & 0.7825 \\
        F1-Score & 0.9064 & 0.6800 & 0.8385 \\
        \midrule
        \multicolumn{4}{l}{\textit{\textbf{Argument Metrics}}} \\
        Task-ArgName F1 & 0.8848 & 0.3495 & 0.4497 \\
        \midrule
        \multicolumn{4}{l}{\textit{\textbf{Sequence Metrics}}} \\
        Edit Distance ($\downarrow$) & 0.1044 & 0.1910 & 0.1073 \\
        Sequence Similarity ($\uparrow$) & 0.8956 & 0.8090 & 0.8927 \\
        \midrule
        \multicolumn{4}{l}{\textit{\textbf{Task Type \& Count Metrics}}} \\
        Type Accuracy & 0.6056 & 0.8194 & 0.8606 \\
        N-Tools Accuracy & 0.7814 & 0.6927 & 0.7292 \\
        \bottomrule
    \end{tabular}
\end{table}

This design choice enables exact matching of predicted tools and parameters against the ground truth, yielding a stricter and more reproducible evaluation than token-level similarity. Comparing the structured results in Table~\ref{tab:exp1_2} with the semantic-based results in Table~\ref{table:table_BERT} highlights this advantage: the Tool/Node Name metrics and Sequence metrics directly assess correctness of tool selection and execution ordering---the quantities that ultimately determine whether a plan can be executed successfully---whereas Rouge and BERTScore conflate surface-level lexical variation (e.g., paraphrasing, filler words) with genuine errors. For example, \textit{dailylifeapis} achieves a Tool Name F1 of 0.9064 and Sequence Similarity of 0.8956 under structured evaluation, compared to a BERT~F1 of 0.535 and Rouge-L of 0.504 under semantic evaluation, suggesting that the semantic metrics substantially underestimate actual plan quality due to their sensitivity to superficial text differences.

\subsection{Integrating the Critique into the Planning Loop}
\begin{figure}[H]
    \centering
    \includegraphics[width=0.9\textwidth]{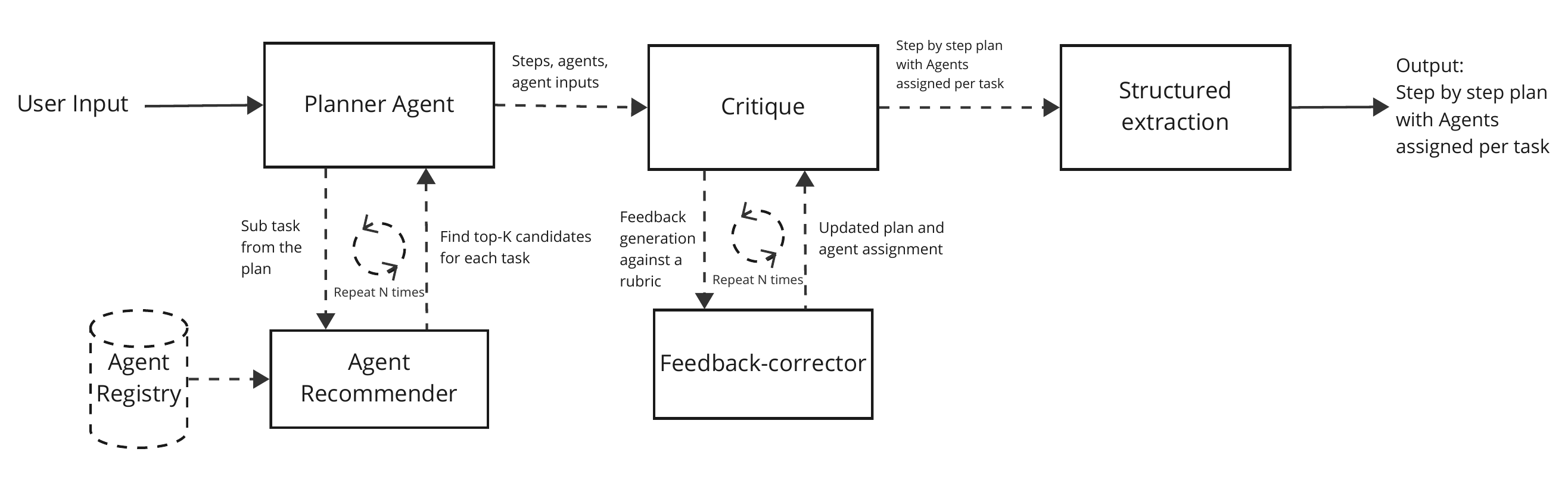}
    \caption{ReAct Style Planning architecture with integrated critique loop. The agent node generates task decompositions and queries the agent recommender iteratively. Upon completion, the extraction stage produces structured predictions which are evaluated by the critique node. When deficiencies are identified and the iteration budget permits, the feedback node provides targeted corrections that preserve valid aspects while addressing identified issues.}
    \label{fig:critic_loop}
\end{figure}

Building upon the critique mechanism introduced in Section \ref{sec:AR}, we conducted an experiment to evaluate the impact of incorporating iterative self-correction into the end-to-end planning and agent recommendation pipeline. The architecture, illustrated in Figure \ref{fig:critic_loop}, augments the basic ReAct-style planner with a critique stage that evaluates generated plans against multiple quality rubrics before finalizing the output.

The critique node operates as a quality gate that evaluates the extracted structured output along five dimensions: (i) logical coherence of the task decomposition, (ii) correctness of identified task nodes, (iii) agreement between predicted tool/agent count and actual tool/agent assignments, (iv) accuracy of workflow type classification, and (v) validity of inter-task dependencies. When the critique identifies deficiencies in any dimension, and the system has not exceeded its iteration budget, control passes to a feedback node that generates targeted corrections. Crucially, this feedback mechanism preserves aspects of the plan that passed validation while addressing only the identified shortcomings, avoiding unnecessary perturbation of correct predictions.

The iterative refinement process continues until either the critique approves the generated plan across all rubrics or a maximum iteration count is reached. This bounded iteration strategy balances the benefits of self-correction against computational cost and the risk of oscillation between alternative valid plans.

Table \ref{tab:exp1_4} presents results with the critique-enhanced architecture. Compared to the baseline results in Table \ref{tab:exp1_2}, the addition of critique and iterative correction yields consistent improvements across nearly all metrics. Tool/agent identification F1-score increases from 0.9064 to 0.9161 for \textit{dailylifeapis}, from 0.6800 to 0.6872 for \textit{huggingface}, and from 0.8385 to 0.8455 for \textit{multimedia}. More notably, sequence similarity improves across all categories, with \textit{dailylifeapis} rising from 0.8956 to 0.9034, indicating that the critique mechanism effectively identifies and corrects ordering errors in task sequences. The type and N-tools accuracy metrics also show improvements, suggesting that the critique's explicit evaluation of workflow structure helps the system converge toward more accurate high-level plan characteristics. These findings corroborate with  our earlier observations regarding the beneficial value of critique in agent recommendation (Section \ref{sec:AR}). demonstrating that LLM based critique evaluation provides complementary benefits when applied to the broader planning and composition task.

However, the mechanism with which the critique can improve the query for a better retrieval is not fully understood and will be further investigated in future work. Our proposed approach is in contrast with previously proposed frameworks (e.g., \cite{shen2024taskbench}) where the complete list of tools is available as context to an LLM, and the LLM can act based on the critique's feedback and improve the selection. In our current approach, the list of candidate agents retrieved by the agent recommender is fixed, and while the critique may alter their order, it cannot alter the selection of the agents. Hence, the critique is not able to improve our reported results any further than what we report here.  However, our approach has substantial token cost and speed advantages compared to previous proposed frameworks such as \cite{shen2024taskbench}, and is scalable to a much larger number of tools/agents.

\begin{table}[h!]
    \centering
    \caption{Metrics for ReAct Style with critique and Iterative Correction. $\uparrow$ higher is better, $\downarrow$ lower is better.}
    \label{tab:exp1_4}
    \begin{tabular}{lccc}
        \toprule
        \textbf{Metric} & \textbf{Dailylife APIs} & \textbf{Hugging Face APIs}& \textbf{Multimedia APIs}\\
        \midrule
        \multicolumn{4}{l}{\textit{\textbf{Tool/Node Name Metrics}}} \\
        Precision & 0.9501 & 0.7497 & 0.9112 \\
        Recall & 0.8845 & 0.6344 & 0.7888 \\
        F1-Score & 0.9161 & 0.6872 & 0.8455 \\
        \midrule
        \multicolumn{4}{l}{\textit{\textbf{Argument Metrics}}} \\
        Task-ArgName F1 & 0.8913 & 0.3547 & 0.4524 \\
        \midrule
        \multicolumn{4}{l}{\textit{\textbf{Sequence Metrics}}} \\
        Edit Distance ($\downarrow$) & 0.1031 & 0.1859 & 0.1051 \\
        Sequence Similarity ($\uparrow$) & 0.9034 & 0.8168 & 0.9012 \\
        \midrule
        \multicolumn{4}{l}{\textit{\textbf{Task Type \& Count Metrics}}} \\
        Type Accuracy & 0.6212 & 0.8277 & 0.8752 \\
        N-Tools Accuracy & 0.7954 & 0.7089 & 0.7401 \\
        \bottomrule
    \end{tabular}
\end{table}

\section{Conclusion and Future Work}
This work was motivated by the need to automate the creation of multi-agent systems that convert user intents into functioning applications and solutions with minimal human intervention. Our proposed an ML-based agentic framework adapts, makes decisions, and changes the course of its actions based on user intent, user requirements, environmental factors, and intermediate results. We introduce an agent recommender that provides an effective solution for navigating the growing complexity of agent registries that may hold thousands of entries. Its two-stage architecture is augmented by the use of  critique mechanism and the enrichment of agent descriptions. Our architecture demonstrates  superior performance in identifying and ranking suitable agents for diverse tasks. The experimental findings robustly support the hypothesis that a multi-stage recommender, particularly with LLM-based re-rankers and agent description enrichment, significantly enhances agent selection. The re-ranker's ability to boost early precision is crucial for user satisfaction in agent-driven applications. However, challenges persist. One notable observation is that LLM-based re-rankers have limited instruction-following capability, exhibiting a bias towards lower cost agents even when explicitly instructed to prefer higher cost ones. This suggests that LLMs might carry inherent biases from their pre-training, requiring careful prompt engineering or fine-tuning for specific optimization criteria. Decoupling ingestion and querying processes, as planned, will further enhance system flexibility and scalability.

Some of the results reported in previous articles on TaskBench show better performance in some categories by providing an entire list of tools as context to the LLM. We believe that the aforementioned approach of providing LLMs with the context and rationale to pick the correct set of agents out of, say, 200 agents (e.g., TaskBench has under 200 tools) will not scale when the number of agents grows dramatically. This is due to increased token cost, limitation in context lengths, processing latency, and LLMs' inability to retain context over large token lengths. Our approach, using retrieval, however, does scale due to its two-stage architecture that uses a fast retriever followed by a re-ranker.

\section*{Acknowledgement}
This research was supported by Cisco Outshift, the innovation engine of Cisco Systems, Inc. We'd like to extend our thanks to Aditya Patel, from Cisco Outshift, for his contributions in the early stages of this work. We also thank our Cisco Outshift colleagues, Ramana Kompalla, Peter Bosch, Ali Payani, Charles Fleming, and Vijoy Pandey, for reviewing this paper and providing valuable comments.

\bibliographystyle{plain}
\nocite{*}
\bibliography{reference}

\end{document}